\documentclass[conference]{IEEEtran}
\IEEEoverridecommandlockouts
\usepackage{cite}
\usepackage{amsmath,amssymb,amsfonts}
\usepackage{multirow}
\usepackage{algorithmic}
\usepackage{graphicx}
\usepackage{textcomp}
\usepackage{float}
\usepackage[dvipsnames]{xcolor}
\usepackage{caption}
\usepackage{hyperref}
\usepackage[utf8x]{inputenc}
\def\BibTeX{{\rm B\kern-.05em{\sc i\kern-.025em b}\kern-.08em
    T\kern-.1667em\lower.7ex\hbox{E}\kern-.125emX}}

\begin{document}

\title{Sim2Real for Peg-Hole Insertion with Eye-in-Hand Camera\\}


\author{\IEEEauthorblockN{Fedor Chervinskii\textsuperscript{\textsection},
Alexander Rybnikov\textsuperscript{\textsection},
Damian Bogunowicz\textsuperscript{\textsection}
and
Komal Vendidandi\textsuperscript{\textsection}}
\IEEEauthorblockA{$\Lambda$ $\Gamma$ $\Gamma$ I V $\Lambda$ L}}

\maketitle
\begingroup\renewcommand\thefootnote{\textsection}
\footnotetext{Equal contribution. \newline [chervinskii, rybnikov, bogunowicz, vendidandi]@arrival.com}
\endgroup


\begin{abstract}
Even though the peg-hole insertion is one of the well-studied problems in robotics, it still remains a challenge for robots, especially when it comes to flexibility and the ability to generalize. Successful completion of the task requires combining several modalities to cope with the complexity of the real world. In our work, we focus on the visual aspect of the problem and employ the strategy of learning an insertion task in a simulator. We use Deep Reinforcement Learning to learn the policy end-to-end and then transfer the learned model to the real robot, without any additional fine-tuning.
We show that the transferred policy, which only takes RGB-D and joint information (proprioception) can perform well on the real robot.
\end{abstract}


\section{Introduction}
A successful peg-hole insertion includes a series of sub-tasks like aligning the peg w.r.t hole and inserting in the hole. The task involves concurrent understanding of vision, depth and haptic modalities. This has been shown to pose significant difficulties for the robots. \cite{b1} claims that having a single modality, like vision, does not solve the peg-hole insertion problem accurately and precisely. This is the motivation behind our multi-modal perception system. In this work focus on using vision and optionally proprioception for robotic control, as well as trying to understand precision limits of a trainable vision-based controller.


Reinforcement Learning (RL) allows learning control policies that are difficult to model explicitly. These policies can generalise with respect to the geometry of the peg and the hole, as stated in \cite{b1}.

A UR5e robot used in our work is trained in the CoppeliaSim \cite{coppeliaSim} to learn a peg-hole insertion policy. We are using different model-free RL algorithms embedded in the Catalyst \cite{Catalyst} framework. In our experiments TD3\cite{TD3} converged to higher returns than DDPG\cite{DDPG} or SAC\cite{SAC} as shown in Fig. \ref{model_free} in section \ref{experiments}. Our deterministic control policy is learned from a multi-modal representation consisting of RGB-D images and proprioception. The policy generalises to different colours, as well as the 3D position of the peg and the block.

\section{Related Work and Background}  \label{related_work}
      
\cite{RL_and_imitation} uses a hybrid approach of model-free deep RL and imitation learning to achieve zero-shot Sim2Real transfer on six different tasks with a single agent. The method shows that pre-training on expert demonstrations leads to faster training than learning the policy from scratch.
\cite{GraspGAN} learns to refine synthetic images (domain adaptation) to achieve good grasping success rate using less real-world data. \cite{13} trains a network that maps input image to the robot's motor torque. \cite{b1} uses self-supervision to learn a compact and multimodal representation of the sensory inputs for the peg-hole insertion task. While all aforementioned works incorporate vision, none of them is using eye-in-hand approach which we consider much more suitable specifically for RL, where a policy can learn to optimize observations along the way, making the vision "active".

\section{Problem Statement and Method Overview}\label{problem_statement}
The main objective of our work is to make the robot learn a peg-hole insertion policy, which is invariant to the pose and color of the mating part or peg. This is done by training a robot in the simulation and then directly transferring the model to the real robot. For a better Sim2Real transferability, we experiment with image augmentation and include different modalities. We benchmark our approach against a proposed baseline model, introduced in section \ref{models}.

\section{Baseline and RL Models}\label{models}

\subsection{Baseline Model}
Our baseline for the peg-in-hole insertion uses visual feedback together with classical computer vision tools to control the insertion of the peg. Firstly, the robot uses color information to segment out the mating part from the image. This allows to filter all the irrelevant visual noise. Secondly, the centre of the hole is being estimated. Finally, the end effector is guided to minimize the distance between the pose of the peg and the estimated pose of the hole.


\subsection{Reinforcement Learning}

We employ the actor-critic method, where both the actor and the critic are parameterized using neural networks. 
\begin{equation}
R_{total} = R_{distance} + R_{success} + R_{collision} + R_{time}
\label{reward_design}
\end{equation}
The total reward stated in Eq. \ref{reward_design} for each step consists of following components. 
\begin{itemize}
    \item \textbf{Distance Reward:} Based on the distance between the the peg and the target (the hole of the mating part). The smaller the distance, the higher the reward.
    \item \textbf{Success Reward:} A sparse reward based on the distance between the peg and the desired position in the hole. The agent is being rewarded once the peg gets very close to the target (specified by a threshold).
    \item \textbf{Collision Reward:} To discourage the robot from colliding with the block, a negative reward is obtained every time the agent collides with the mating part.
    \item \textbf{Time Reward:} To coerce the agent to complete the task as fast as possible, it obtains a small penalty per every step. 
\end{itemize}

\begin{figure}[ht]
\centering
\includegraphics[width=0.4\textwidth]{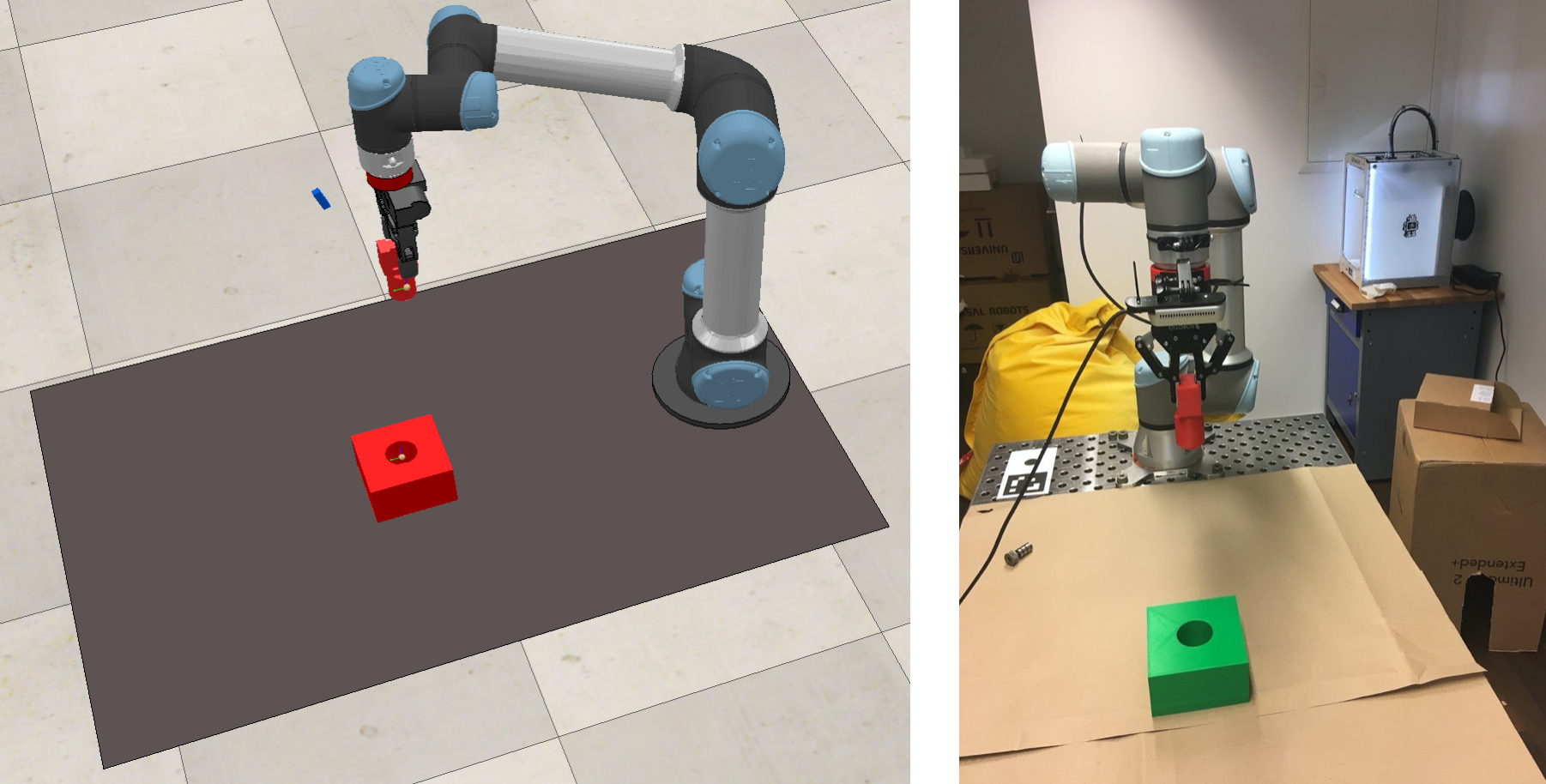}
\caption{\textit{Left:} Rendered simulator scene from CoppeliaSim showing the UR5e robot with Robotiq gripper, a peg and a block. \textit{Right:} Real UR5e robot in our lab.} 
\label{vrep}
\end{figure}

The scene in the CoppeliaSim and the real robot in starting position are shown in Fig. \ref{vrep}. On the real robot we use RealSense D435 RGB-D active stereo sensor. The mating part is being randomly placed on the table (always fully or partially visible for the robot in the starting position), imitating a flexible assembly process.


\section{Experiments: Results} \label{experiments}
\begin{figure}[ht]
\centering
\includegraphics[width=0.45\textwidth]{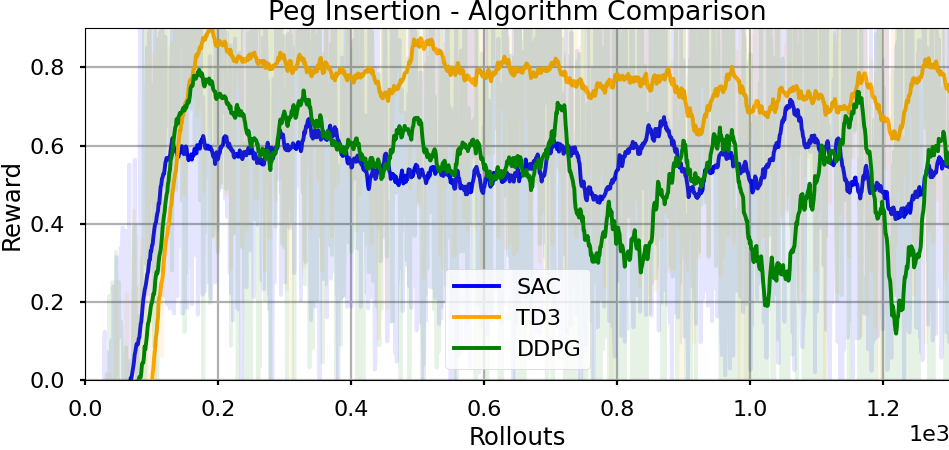}
\caption{ TD3 yielding higher return than SAC and DDPG.}
\label{model_free}
\end{figure}

\begin{figure}[ht]
\centering
\includegraphics[width=0.45\textwidth]{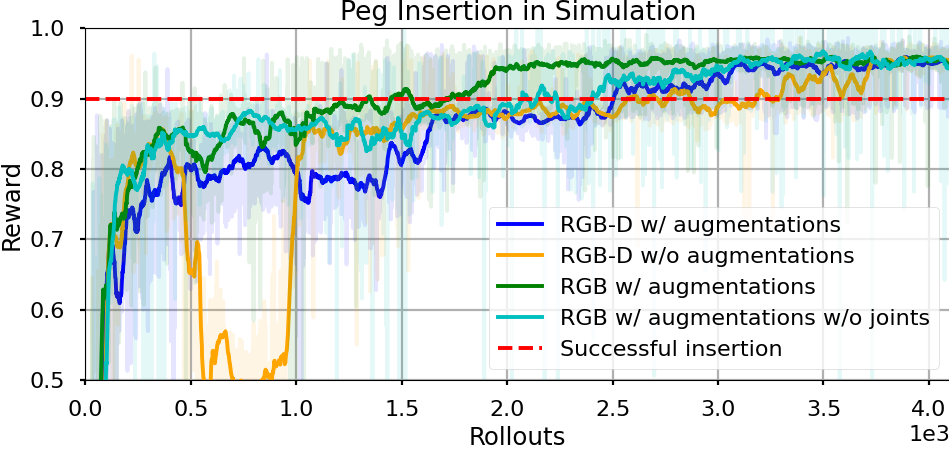}
\caption{Ablation study showing different configurations of training in learning a peg-hole insertion policy in simulation. It can be seen that all the configurations converged to successful insertion in simulation. The performance of these configurations on real robot can be seen in Table \ref{tab:metrics}. } 
\label{ablation}
\end{figure}

First, we train the agent in the simulation. The learned policy is directly transferred to our real UR5e robot. Evaluation of the models is shown in Table \ref{tab:metrics}. We did a total of $30$ rollouts for each model, in which the mating part pose is changed for every 5 rollouts. These 6 mating poses are predefined, with a maximum distance of $39$ cm and $21$ cm between the poses in x- and y-axis respectively (in the plane of the table). The robot starts at a predefined home position for every rollout in both simulation and real world. The comparison of different model-free RL algorithms can be found in Fig. \ref{model_free}. Fig. \ref{ablation} shows which modalities are most important for the performance in the simulation.

\begin{table}[H]
\centering
\begin{tabular}{|c|c|c|c|}
\hline
\multirow{2}{*}{\textbf{Model Type}} & \multicolumn{2}{c|}{\textbf{Mean Error {[}mm{]}}} & \multirow{2}{*}{\textbf{Insertion {[}\%{]}}} \\ \cline{2-3}
                                     & x                       & y                       &                                                 \\ \hline
Baseline                             & \hspace{1mm} $2.4\pm1.7$             & \hspace{0.1mm} $0.3\pm3.0$             & $100$                                           \\ \hline
RGB w/ augs & $-2.0\pm3.8$  &$-9.8\pm11.2$ & $55$                             
\\ \hline
 RGB-D w/ augs & \hspace{1mm} $17.7\pm12.3$  & \hspace{0.1mm} $2.7\pm1.8$ & $0$                                       
\\ \hline
 RGB-D w/o augs  & $-124.1\pm166.9$  & \hspace{0.1mm} $65.6\pm94.4$  & $0$                                      
\\ \hline

\end{tabular}
\caption{\textbf{Mean insertion error and insertion rate on the real robot.} The mean error between the ground truth position of the inserted peg versus the actual position at the end of an episode. Insertion rate is the percentage of the rollouts which end with successful insertion. The mean is taken over $30$ runs for different positions of the mating part on the table.}
\label{tab:metrics}
\end{table}

\section{Discussion and Conclusion}\label{conclusion}
Our initial hypothesis was that image augmentation is crucial for successful transfer from simulation to the real-world. Table \ref{tab:metrics} shows that augmentations improve the precision significantly. However, the RGB-D model still does not achieve steady, successful insertions. That could be because the depth image has a very specific structural noise (depth values are computed from stereo cameras) that can hardly be modelled with standard augmentation algorithms. Thus, RGB-only model shows much better performance. Additionally, the ablation study in Fig. \ref{ablation} shows that proprioceptive features are not critical for model performance for this task, neither in simulation, nor in real-life. Finally, we show that a purely eye-in-hand, image-based controller can be trained in simulation to perform peg-hole insertion with sub-centimetre accuracy. However, we still experience instabilities caused by changing light conditions and other environmental factors, that can be possibly improved by enhancing rendering in simulation and richer augmentation. 
For the future work we consider enhancing the vision-based controller with force-based one (trained end-to-end, as a single model), that can perform precise insertions with large initial uncertainty.




\appendix[Simulation sensor data augmentations]\label{appendix-augs}
Different sets of augmentations are used for the RGB image and the depth image. Augmentations for the latter are stronger due to the significant noise in depth values (computed from RealSense camera stereo vision). \textit{albumentations} \cite{alb} library is used to implement of all augmentations used in this work. RGB image is initially converted to grayscale and then the following augmentations are used: random brightness-contrast, motion blur, Gaussian blur, median blur, solarize filter, embossing, histogram equalization, sharpening, CLAHE, image compression, multiplicative noise, Gaussian noise. For depth images the following augmentations are used: thresholding and dilation, sharpening, embossing, motion blur, Gaussian blur, median blur, multiplicative noise, Gaussian noise, coarse "pepper" dropout, coarse "salt" dropout.
\begin{figure}[ht]
\centering
\includegraphics[width=0.45\textwidth]{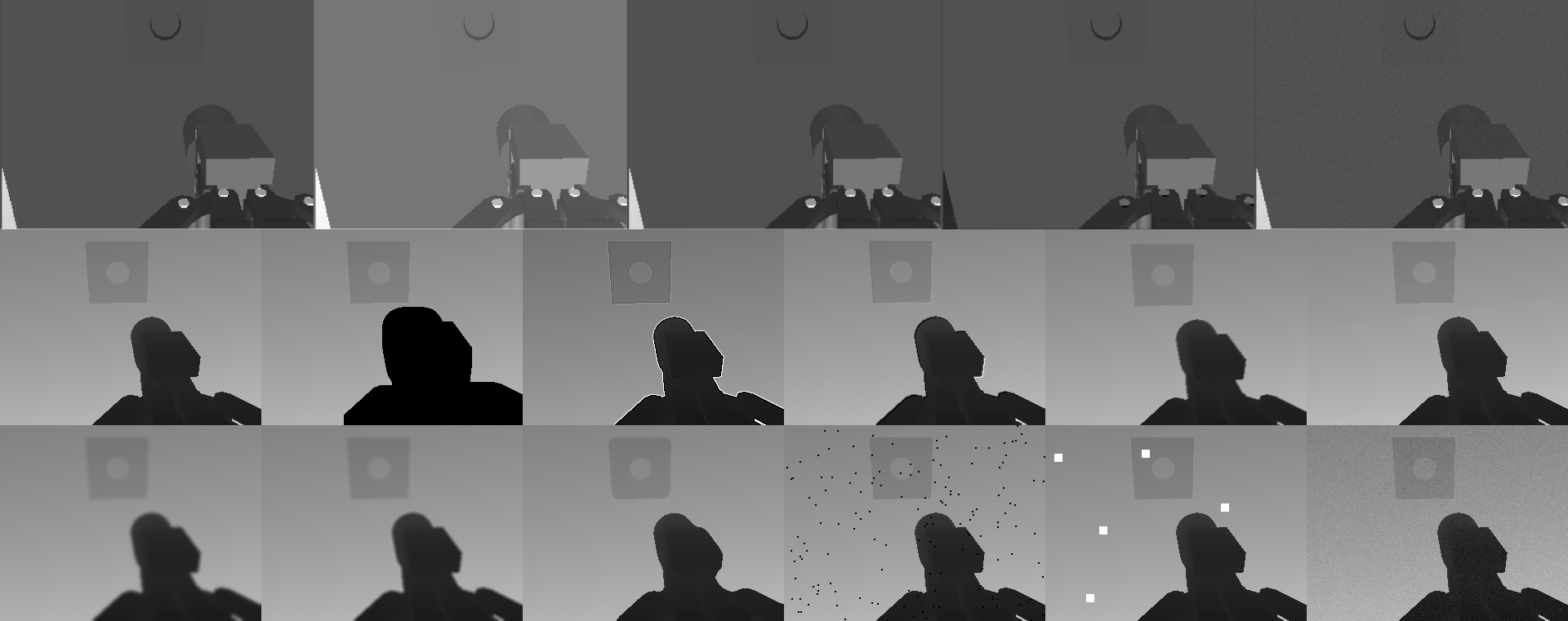}
\caption{Examples of used augmentations. \textit{Top row, left to right:} original grayscale image, random brightness-contrast, motion blur, solarize filter, Gaussian noise. \textit{Middle row, left to right:} original depth image, threshold and dilation, sharpening, embossing, motion blur, multiplicative noise. \textit{Bottom row, left to right:} Gaussian blur, blur, median blur, coarse "pepper" dropout, coarse "salt" dropout, Gaussian noise.}
\label{augs-fig}
\end{figure}


\begin{thebibliography}{00}
\bibitem{b1} Lee, Michelle A., et al. "Making sense of vision and touch: Self-supervised learning of multimodal representations for contact-rich tasks." 2019 International Conference on Robotics and Automation (ICRA). IEEE, 2019.

\bibitem{coppeliaSim} E. Rohmer, S. P. N. Singh, M. Freese, "CoppeliaSim (formerly V-REP): a Versatile and Scalable Robot Simulation Framework", IEEE/RSJ Int. Conf. on Intelligent Robots and Systems, 2013. www.coppeliarobotics.com

\bibitem{Catalyst} Kolesnikov, Sergey, and Oleksii Hrinchuk. "Catalyst. RL: A Distributed Framework for Reproducible RL Research." arXiv preprint arXiv:1903.00027 (2019).

\bibitem{TD3} Fujimoto, Scott, Herke Van Hoof, and David Meger. "Addressing function approximation error in actor-critic methods." arXiv preprint arXiv:1802.09477 (2018). 

\bibitem{DDPG} Silver, David, et al. "Deterministic policy gradient algorithms." 2014.

\bibitem{SAC} Haarnoja, Tuomas, et al. "Soft actor-critic: Off-policy maximum entropy deep reinforcement learning with a stochastic actor." arXiv preprint arXiv:1801.01290 (2018).

\bibitem{RL_and_imitation} Zhu, Yuke, et al. "Reinforcement and imitation learning for diverse visuomotor skills." arXiv preprint arXiv:1802.09564 (2018).

\bibitem{GraspGAN} Bousmalis, Konstantinos, et al. "Using simulation and domain adaptation to improve efficiency of deep robotic grasping." 2018 IEEE International Conference on Robotics and Automation (ICRA). IEEE, 2018.

\bibitem{13} Levine, Sergey, et al. "End-to-end training of deep visuomotor policies." The Journal of Machine Learning Research 17.1 (2016): 1334-1373.

\bibitem{alb} \textit{albumentations} python module \url{https://albumentations.readthedocs.io/en/latest/api/augmentations.html}
\end{thebibliography}
\end{document}